\documentclass[conference]{IEEEtran}
\IEEEoverridecommandlockouts
\usepackage{cite}
\usepackage{amsmath,amssymb,amsfonts}
\usepackage{graphicx}
\usepackage{textcomp}
\usepackage{xcolor}
\usepackage{booktabs}    
\usepackage{colortbl}    
\usepackage{booktabs}    
\usepackage{colortbl}    
\usepackage{xcolor}      
\usepackage{multirow}    
\usepackage{adjustbox}   
\usepackage[ruled,vlined]{algorithm2e}
\usepackage{hyperref}

\def\BibTeX{{\rm B\kern-.05em{\sc i\kern-.025em b}\kern-.08em
    T\kern-.1667em\lower.7ex\hbox{E}\kern-.125emX}}
\begin{document}

\title{Conditional Consistency Guided Image Translation and Enhancement}

\author{
\IEEEauthorblockN{Amil Bhagat, Milind Jain, A. V. Subramanyam}
\IEEEauthorblockA{\textit{Indraprastha Institute of Information Technology (IIIT)}, Delhi, India \\
Email: \{amil21309@iiitd.ac.in, milind21165@iiitd.ac.in, subramanyam@iiitd.ac.in\}}
}

\maketitle

\maketitle

\begin{abstract}
Consistency models have emerged as a promising alternative to diffusion models, offering high-quality generative capabilities through single-step sample generation. However, their application to multi-domain image translation tasks, such as cross-modal translation and low-light image enhancement remains largely unexplored. In this paper, we introduce Conditional Consistency Models (CCMs) for multi-domain image translation by incorporating additional conditional inputs. We implement these modifications by introducing task-specific conditional inputs that guide the denoising process, ensuring that the generated outputs retain structural and contextual information from the corresponding input domain. We evaluate CCMs on 10 different datasets demonstrating their effectiveness in producing high-quality translated images across multiple domains. Code is available \url{https://github.com/amilbhagat/Conditional-Consistency-Models}.
\end{abstract}


\section{Introduction}
\label{sec:intro}

Generative modelling has witnessed rapid progress in recent years, evolving from the early successes of Generative Adversarial Networks (GANs) \cite{goodfellow2014generative} to the more stable and controllable diffusion models \cite{ho2020denoising}. In the early stages, the introduction of conditional GANs (cGANs) \cite{mirza2014conditional} made it possible to guide the generation process using auxiliary information such as class labels or paired images. This allowed models to learn structured mappings between input and target domains in a variety of conditional image-to-image translation tasks, including semantic segmentation, image synthesis, style transfer and other applications \cite{isola2017image, zhu2017unpaired}. Despite their successes, conditional GANs often suffer from mode collapse, training instability, and the necessity of intricate adversarial training schemes.

On the other hand, score-based diffusion models \cite{ho2020denoising} have emerged as a robust alternative for image generation and translation. These models add noise to the data and learn to invert this process via score estimation, producing high-quality outputs. However, these models require many iterative steps to produce high quality images. Recently, consistency models \cite{song2023consistency} have been proposed as a new class of generative models. Instead of relying on iterative refinement steps like diffusion, consistency models directly learn mappings from any noise level to data. 

However, despite their advantages, consistency models have been explored primarily in unconditional settings. The extension of these models to conditional, multi-domain tasks remains under-explored. This paper focuses on two representative tasks of significant practical value and complexity: cross-modal translation and low-light image enhancement (LLIE).

Cross-modal translation is a crucial task in surveillance and medical systems where different modalities may reveal complementary information. The LLVIP dataset \cite{zhang2021llvip} exemplifies the significance of this translation, as methods must effectively learn visible to infrared mappings while preserving the spatial details. Similarly, BCI \cite{liu2022bci} provides a challenging medical dataset of paired hematoxylin and eosin (HE) and immunohistochemical techniques (IHC) images. As routine diagnosis with IHC is very expensive, translation task from HE to IHC can be very helpful in bringing down the overall cost.
On the other hand, LLIE is critical in computer vision applications where images suffer from insufficient illumination, leading to poor image details. Methods such as SID \cite{chen2018learning}, GSAD \cite{hou2024global} have demonstrated impressive results using data-driven approaches trained on paired dark and well-exposed images. Nonetheless, these methods often rely on adversarial training setups which are known to have convergence issues, or explicit iterative sampling strategy which is time consuming.

While existing approaches to these problems range from traditional Retinex-based methods \cite{land1971lightness} to deep convolutional neural networks \cite{chen2018learning}, many either lack the adaptability required for multi-domain translation or are computationally expensive at inference time. Conditional diffusion models can produce excellent results but at the cost of significant computational overhead during sampling \cite{saharia2022image}. Conditional GANs, though fast, may exhibit mode collapse or require carefully tuned adversarial training setups \cite{che2016mode}.

In this work, we introduce Conditional Consistency Models (CCMs) for multi-domain image translation tasks. CCMs innovate on traditional consistency models by incorporating additional conditional inputs, such as, visible image for visible to infrared translation, HE image for HE to IHC translation, and low-light image for LLIE, respectively. By carefully designing the network architecture to take the conditional input and training process, CCMs yield highly efficient inference while preserving important local and global spatial details from the conditional input. Unlike other techniques, our method is applicable for both translation and image enhancement.

\section{Related Works}

\subsection{Cross-modal Image Translation}
Generative approaches for image translation have evolved significantly. In case of visible to infrared translation, GAN-based methods have been widely explored, and models such as ThermalGAN \cite{kniaz2018thermalgan}, TMGAN \cite{ma2024visible}, and InfraGAN \cite{ozkanouglu2022infragan} have achieved notable success in generating high-quality IR images. These methods often require paired RGB-IR datasets, which are scarce in practice. Lee \textit{et. al.} propose an unsupervised method using edge-guidance and style control in a mutli-domain image translation framework. MappingFormer \cite{wang2024mappingformer} uses low and high frequency features for visible to infrared image generation. These features are fused and a dual contrast learning module further finds an effective mapping between the unpaired data. Similarly, \cite{zhu2024data} also study the unpaired data translation.


In case of medical images, BCI \cite{liu2022bci} addresses translation of HE images to IHC images. They propose a multi-scale structure in a Pix2Pix framework. The multi-scale pyramid enables the model to capture both global and local features effectively leading to improved image generation quality. On similar lines, GAN based translation is investigated in \cite{li2023adaptivesupervisedpatchnceloss, zhang2024high, chen2024pathological}.

\subsection{Low Light Image Enhancement}

\subsubsection*{GAN Based Methods} 
EnlightenGAN \cite{jiang2021enlightengan} adopts a global and local discriminator to directly map low-light images to normal-light images. This is trained in unsupervised manner. 
Similarly, other unsupervised methods show further improvement in the enhanced images \cite{yang2023implicit}.
 
 \subsubsection*{Diffusion Based Methods} Diff-Retinex \cite{yi2023diff} combines Retinex decomposition with diffusion-based conditional generation. 
 It separates images into illumination and reflectance components, which are then adjusted using diffusion networks. 
 ExposureDiffusion \cite{wang2023exposurediffusion} integrates physics-based exposure processes with a diffusion model. Instead of starting from pure noise, it begins with noisy images. It uses adaptive residual layers to refine image details and achieves better generalization and performance with fewer parameters. AnlightenDiff \cite{chan2024anlightendiff} formulates the problem of LLIE as residual learning. It decomposes the difference between normal and low light image into a residual component and a noise term. PyDiff \cite{zhou2023pyramid} introduces a pyramid resolution sampling strategy to speed up the diffusion process and employs a global corrector to address degradation issues. GSAD \cite{hou2024global} uses a curvature-regularized trajectory in the diffusion process to preserve fine details and enhance global contrast. It also incorporates uncertainty-guided regularization to adaptively focus on challenging regions. These methods use paired datasets.


\subsection{Preliminaries: Consistency Models}



Consider a continuous trajectory \(\{\mathbf{r}_t\}_{t \in [\epsilon, T]}\) connecting a clean data sample \(\mathbf{r}_\epsilon\) at time \(\epsilon\) to a noisy sample \(\mathbf{r}_T\) at a larger time \(T\). The probability flow ODE ensures a bijective mapping between data and noise at different time scales. A consistency model \cite{song2023consistency} leverages this structure to learn a direct, single-step mapping from a noisy sample at any time \(t\) back to the clean sample \(\mathbf{r}_\epsilon\). Formally, a consistency model defines a consistency function \(g_\phi\) such that:
\begin{equation}
g_\phi(\mathbf{r}_t, t) = \mathbf{r}_\epsilon \quad \forall \; t \in [\epsilon, T].
\label{eq:consistency_condition_all_t}
\end{equation}
This \emph{self-consistency} property ensures that no matter the noise level \(t\), the model consistently recovers the same underlying clean data sample.

\subsubsection*{Boundary Condition and Parameterization}

A key requirement is the \emph{boundary condition}, stating that at the minimal noise level \(\epsilon\), the model should act as the identity 
$g_\phi(\mathbf{r}_\epsilon, \epsilon) = \mathbf{r}_\epsilon$. 
To naturally incorporate this condition, consistency models employ a parameterization that respects this constraint. A common approach uses a combination of skip connections and time-dependent scaling:
\begin{equation}
g_\phi(\mathbf{r}, t) = a_{\text{skip}}(t)\,\mathbf{r} + a_{\text{out}}(t)\,G_\phi(\mathbf{r}, t),
\end{equation}
where, $a_{\text{skip}}(t)$ and $a_{\text{out}}(t)$ are scalar functions of $t$ that regulate the contributions of the input $\mathbf{r}$ and the learned function $G_\phi(\mathbf{r}, t)$, respectively. The boundary condition is satisfied by setting $a_{\text{skip}}(\epsilon) = 1$ and $a_{\text{out}}(\epsilon) = 0$, ensuring that $g_\phi(\mathbf{r}, \epsilon) = \mathbf{r}$.

As the noise scale \(t\) increases, the influence of \(G_\phi(\mathbf{r}, t)\) grows through \(a_{\text{out}}(t)\), allowing the model to move away from the identity mapping and perform increasingly complex denoising transformations. This parameterization naturally embeds the boundary condition in the model structure and maintains differentiability across noise scales.

\subsubsection*{Training Consistency Models}

Training a consistency model involves enforcing self-consistency at multiple noise levels. Given a sample \(\mathbf{r}\), a noise vector \(\mathbf{z} \sim \mathcal{N}(\mathbf{0}, \mathbf{I})\), and a pair of noise scales \((t_n, t_{n+1}) \in [\epsilon, T]\), we form noisy inputs 
$\mathbf{r}_{t_n} = \mathbf{r} + {t_n}\mathbf{z}, \quad \mathbf{r}_{t_{n+1}} = \mathbf{r}+ {t_{n+1}}\mathbf{z}$.
The model \(g_\phi\) should produce the same clean output \(\mathbf{r}_\epsilon\) for both \(\mathbf{r}_{t_n}\) and \(\mathbf{r}_{t_{n+1}}\). Thus, the training loss encourages:
\[
g_\phi(\mathbf{r}_{t_{n+1}}, t_{n+1}) = g_\phi(\mathbf{r}_{t_n}, t_n) = \mathbf{r}_\epsilon.
\]
By minimizing a suitable distance measure between these outputs, the network learns to invert the noise injection step at any arbitrary noise level \(t\).

\section{Proposed Method}
\label{sec:proposed_method}

\begin{figure}[h!]
    \centering
  
    \includegraphics[width=\linewidth]{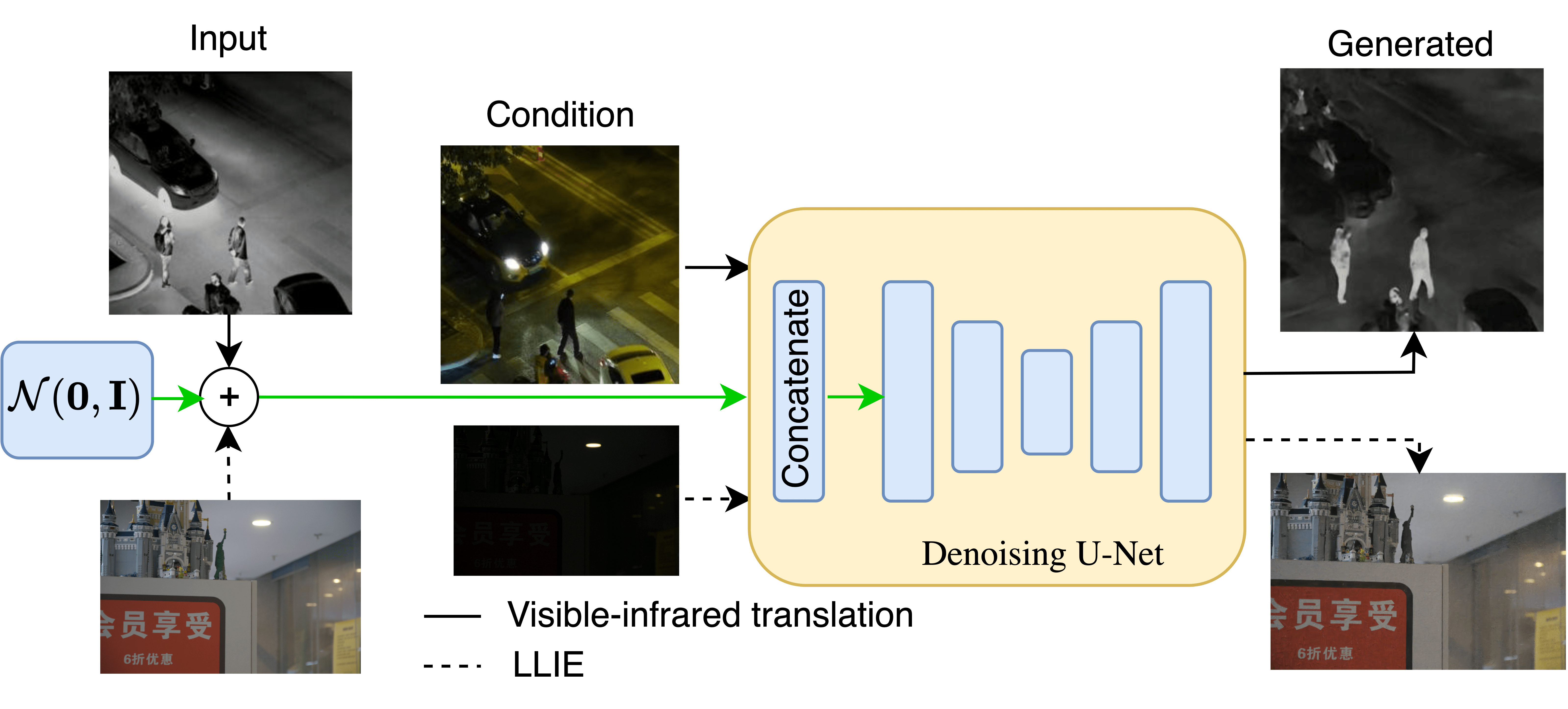}
    \caption{Model Architecture. Our model can take a pair of visible-infrared, HE-IHC, or, low light and well-exposed images. Visible, HE or low light image acts as a conditional input. The noise as per time step $t$ is added to the input infrared, IHC or well exposed image. This noisy image is then concatenated with the condition input and fed to the U-Net. The model can be sampled to obtain the infrared, IHC or enhanced image.}
    \label{fig:example}
\end{figure}

Given a paired dataset $\mathcal{D}=\{\mathbf{v}_i,\mathbf{r}_i\}_{i=1}^N$ of RGB $\mathbf{v}_i \in \mathbb{R}^{H \times W \times C}$ and its corresponding  pair $\mathbf{r}_i\in \mathbb{R}^{H \times W \times C}$. In case of visible to infrared image translation, $\mathbf{v}$ is the RGB image and $\mathbf{r}$ is the infrared image. For BCI dataset \cite{liu2022bci}, $\mathbf{v}$ is the HE image and $\mathbf{r}$ is the IHC image. In case of LLIE, $\mathbf{v}$ is the ill-exposed image and $\mathbf{r}$ is the well-exposed image. Our aim is to learn a mapping from a given input and condition image to its respective paired image. More formally, we learn a function $G:\mathbf{v}\times \mathbf{r} \rightarrow \mathbf{r}$, where $\mathbf{v}\times \mathbf{r}$ denotes a pair. In our proposed method, $G$ is parameterized via a consistency model \cite{song2023improved}. In order to generate images which semantically align with the input image, we make use of conditional synthesis. Here, we use an image pair ($\mathbf{v}, \mathbf{r}$) as the condition and input, respectively, and train the model to generate only $\mathbf{r}$. Figure \ref{fig:example} illustrates the proposed method.

\subsection{Conditional Consistency}

To adapt consistency models to conditional multi-domain image translation, we introduce a conditional input \(\mathbf{v}\). We define the conditional consistency function as,

\begin{equation}
g_\phi(\mathbf{r}, \mathbf{v}, t) = 
\begin{cases}
 \mathbf{r}& t = \epsilon, \\
G_\phi(\mathbf{r}, \mathbf{v}, t), & t\in (\epsilon, T],
\end{cases}
\label{eq:conditional_consistency_equation}
\end{equation}

where \(\mathbf{v}\) provides conditional information that guides the denoising process. As \(g_\phi\) inverts the noise injection step, it also conditions on \(\mathbf{v}\) to ensure that the generated image \(\mathbf{r}\) corresponds to the infra-red, IHC, or well-exposed pair of \(\mathbf{v}\).

In addition, the consistency function must follow the boundary condition, that is:
\begin{equation}
    g_\phi(\mathbf{r}_\epsilon, \mathbf{v}, \epsilon) = \mathbf{r}_\epsilon.
    \label{eq:boundary}
\end{equation}
As the consistency function must map noisy data from any time-step to the clean data,
 using eq \ref{eq:conditional_consistency_equation} and eq \ref{eq:boundary}, we can write,
\[
g_\phi(\mathbf{r}_{t_{n+1}}, \mathbf{v}, t_{n+1}) = g_\phi(\mathbf{r}_{t_n}, \mathbf{v}, t_n) = \mathbf{r}_\epsilon.
\]


We parameterize the conditional consistency model as,
 \begin{equation}
g_\phi(\mathbf{r},\mathbf{v}, t) = a_{\text{skip}}(t) \mathbf{r} + a_{\text{out}}(t) G_\phi(\mathbf{r}, \mathbf{v}, t).
\label{eq:gphi_skip}
\end{equation}

In eq \ref{eq:gphi_skip}, the boundary condition, where \( g_\phi(\mathbf{r, v}, t) = \mathbf{r} \), is obtained for \( t = \epsilon \), $a_{\text{skip}}(\epsilon) = 1$ and $a_{\text{out}}(\epsilon) = 0$. This ensures that the model outputs the clean  image \( \mathbf{r} \) at the minimal noise scale. 
 As the noise scale increases, the generative component \( G_\phi(\mathbf{r, v}, t) \) plays a more significant role, allowing the network to learn complex transformations required for translating \( \mathbf{v} \) into \( \mathbf{r} \). We present the training process in Algoirthm \ref{algo:cct}.

\subsection{Training and Sampling}

\subsubsection*{Training Algorithm}
In order to generate the desired output given the inputs, we make use of conditional synthesis. Here, we use an image pair ($\mathbf{v}, \mathbf{r}$) as the input and train the model to generate only $\mathbf{r}$. $G_{\phi}$ takes a 2$C$ channel input and gives a $C$ channel output. In our method, we concatenate $\mathbf{v}$ and $\mathbf{r}$ across channels to obtain a $2C$ channel input. $\mathbf{v}$ only acts as a conditional input and we do not add any noise to it.

\begin{algorithm}
\caption{Conditional Consistency Training (CCT)}
\label{algo:cct}
\KwIn{Paired dataset $\mathcal{D} = \{(\mathbf{v}_i, \mathbf{r}_i)\}_{i=1}^N$, initial model parameter $\phi$, learning rate $\eta$, step schedule $M(\cdot)$, distance function $d(\cdot, \cdot)$, weighting function $\lambda(\cdot)$}
Initialize $\phi^- \leftarrow \phi$, $k \leftarrow 0$\;
\Repeat{Convergence}{
Sample $(\mathbf{v}, \mathbf{r}) \sim \mathcal{D}$ and $n \sim U[1, M(k) - 1]$\\
Sample $\mathbf{z} \sim \mathcal{N}(\mathbf{0}, \mathbf{I})$\\
    Compute the loss:
    \begin{align*}
    \mathcal{L}(\phi, \phi^-) \leftarrow 
    &\lambda(t_n) d\Big(g_\phi(\mathbf{r} + t_{n+1} \mathbf{z}, \mathbf{v}, t_{n+1}), \nonumber \\
    &\quad g_{\phi^-}(\mathbf{r} + t_n \mathbf{z}, \mathbf{v}, t_n)\Big)
    \end{align*}
    Update the model parameters:
    \[
    \phi \leftarrow \phi - \eta \nabla_\phi \mathcal{L}(\phi, \phi^-)
    \]
    \[
    \phi^- \leftarrow \phi
    \]
    Increment $k \leftarrow k + 1$\
}
\end{algorithm}

We adopt the step schedule $M(.)$ as given in \cite{song2023improved} and the distance function $d(.,.)$ is pseudo-Huber loss given in \cite{song2023improved}.



    
    
    
    
    


\subsubsection*{Sampling }

The unconditional sampling procedure starts from Gaussian noise \(\hat{\mathbf{r}}_T\) and evaluates the consistency model $\mathbf{r} \leftarrow g_\phi(\hat{\mathbf{r}}_T, T)$. To incorporate conditions, we now have:
\begin{equation}
\mathbf{r} \leftarrow g_\phi(\hat{\mathbf{r}}_T, \mathbf{v}, T).
\end{equation}
In practice, we concatenate the $C$ channel noise with the condition input $\mathbf{v}$ and evaluate $g_{\phi}$.
This modified algorithm ensures that sampling is guided by the conditional input \( \mathbf{v} \). We sample through single-step generation only.

\section{Experiment}
In the experiments we show that our method is applicable to different tasks and based on the conditional input, the paired output can be generated. We compare our results with SOTA methods in image translation and LLIE tasks, and demonstrate that our method achieves competitive results. Additionally, we would like to emphasize that our method generalizes well to medical dataset also.

\subsection{Datasets}
We evaluate our proposed method on multiple datasets, including LLVIP \cite{zhang2021llvip}, BCI \cite{liu2022bci}, LOLv1 \cite{wei2018deep}, LOLv2 \cite{yang2021sparse}, and SID \cite{chen2018learning}. The LLVIP dataset comprises 15,488 pairs of visible and thermal images captured under low-light conditions, of which 12,025 pairs are for training and 3,463 are reserved for testing. The BCI dataset comprises 9,746 images, organized into 4,873 pairs of Hematoxylin and Eosin (HE), and Immunohistochemistry (IHC) images. Among these, 3,896 pairs are used for training, while the remaining 977 pairs are used for testing. LOLv1 comprises 485 training-paired low-light and normal-light images and 15 testing pairs. LOLv2 is divided into LOLv2-real and LOLv2-synthetic subsets, each containing 689 and 900 training pairs and 100 testing pairs, respectively. For SID dataset, we used the subset captured by the Sony $\alpha$ 7S II camera. It comprises 2,463 short-/long-exposure RAW image pairs, of these, 1,865 image pairs are used for training, while 598 pairs are used for testing.
In addition to the above benchmarks, we tested our
method on five unpaired datasets, LIME \cite{guo2016lime}, NPE \cite{6512558}, MEF \cite{7120119} and DICM \cite{6615961} and VV\footnote{\url{https://sites.google.com/site/vonikakis/datasets}}, that are low-light images and used to check the LLIE potential of a model.

\subsection{Implementation Details}
LLVIP images are randomly cropped to 512x512 and then resized to 128x128 for training over 1000 epochs. In case of BCI, images are randomly cropped to 256x256 and trained for 500 epochs. LOLv1, LOLv2-real LOLv2-synthetic and SID images are randomly cropped to 128x128 for training over 500 epochs for SID, and 1500 epochs for the rest. The values of $a_{skip}(t)$, $a_{out}(t)$, $\lambda(t_n)$, $M(.)$, and $d(.,.)$ are set to default values given in \cite{song2023improved}.

\subsection{Evaluation Metrics}
We utilize Peak Signal-to-Noise Ratio (PSNR) and Structural Similarity Index Measure (SSIM) to assess the quality of the generated images. Additionally, for datasets lacking paired data, we employ the Naturalness Image Quality Evaluator (NIQE) \cite{mittal2013niqe}.

\section{Results}

\subsection{LLVIP and BCI Results}
\label{sec:llvip_results}
\begin{figure}[h] 
    \centering
    \setlength{\tabcolsep}{2pt} 
    \renewcommand{\arraystretch}{1} 
    \begin{tabular}{ccc} 
        \includegraphics[width=0.15\textwidth]{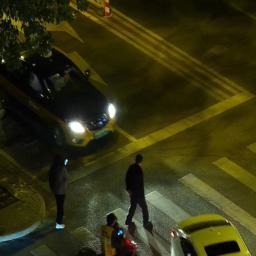} &
        \includegraphics[width=0.15\textwidth]{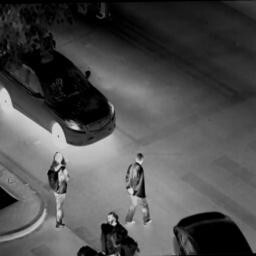} &
        \includegraphics[width=0.15\textwidth]{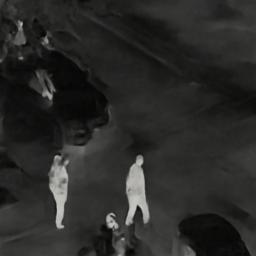} \\
        
        \includegraphics[width=0.15\textwidth]{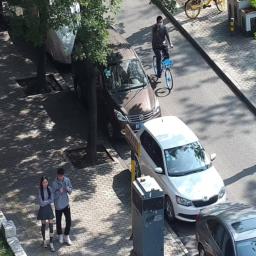} &
        \includegraphics[width=0.15\textwidth]{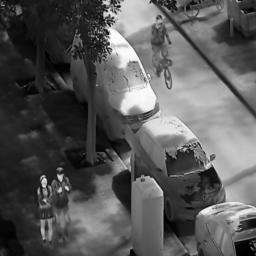} &
        \includegraphics[width=0.15\textwidth]{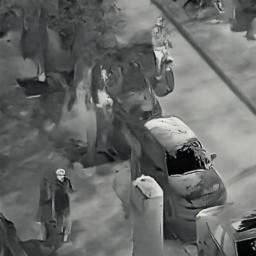} \\
        (a) & (b) & (c) \\
    \end{tabular}
    \caption{Comparison of (a) Visible, (b) Ground Truth Infrared, and (c) Generated Infrared images.}
    \label{fig:vir-comparison}
\end{figure}

We evaluate the performance on the LLVIP dataset in two different ways. The image is either randomly cropped into 512x512, or the full image is resized to 256x256 pixels. We present the results in Table \ref{tab:llvip_comparison}. Our model outperforms other methods. We also show the qualitative results in Fig. \ref{fig:vir-comparison}. It can be observed that different objects are well represented with slight deterioration in spatial details.

\begin{table}[h]
\centering
\caption{Results on LLVIP. Test image resolution is 512$\times$512. (*) represents evaluation where the full image was resized to 256$\times$256. Higher values indicate better performance. Best method is highlighted in bold font.}


\label{tab:llvip_comparison}
\begin{tabular}{l|c|c}
\toprule
Methods & PSNR (dB) & SSIM \\
\midrule
CycleGAN \cite{liu2022bci} & 11.22 & 0.214 \\
BCI \cite{liu2022bci}  & 12.19 & 0.270 \\
pix2pixGAN* \cite{zhang2021llvip}  & 10.76 & 0.175 \\
\midrule
 \textbf{Ours} & \textbf{13.11} & \textbf{0.59} \\
  \textbf{Ours*} & \textbf{12.59} & \textbf{0.49} \\
\midrule

\end{tabular}
\end{table}

\begin{figure}[!ht] 
    \centering
    \setlength{\tabcolsep}{2pt} 
    \renewcommand{\arraystretch}{1} 
    \begin{tabular}{ccc} 
        \includegraphics[width=0.15\textwidth]{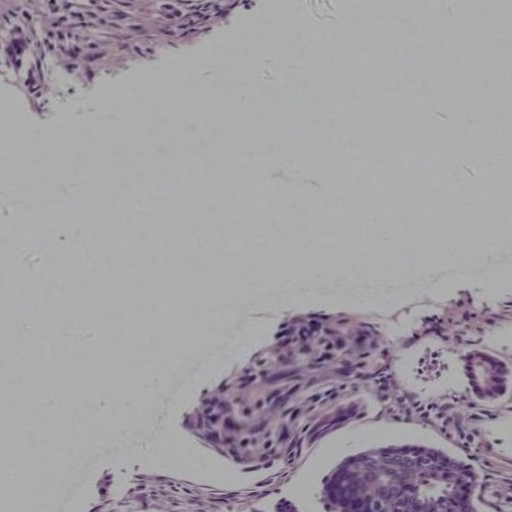} &
        \includegraphics[width=0.15\textwidth]{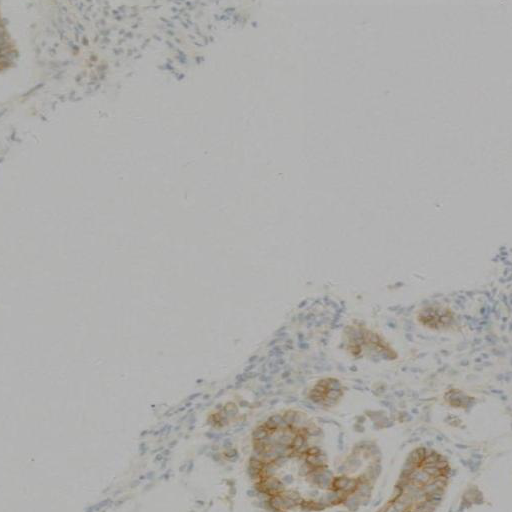} &
        \includegraphics[width=0.15\textwidth]{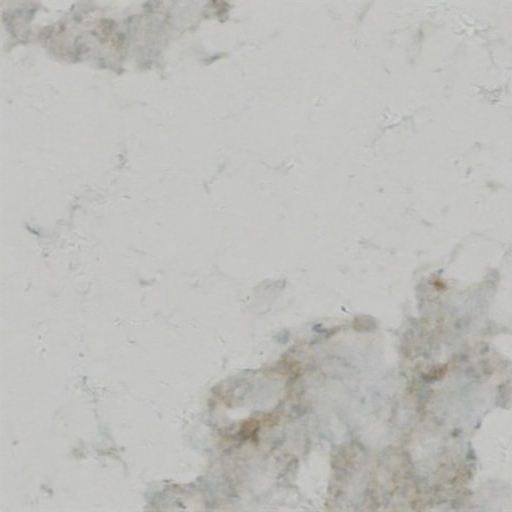} \\
        
        \includegraphics[width=0.15\textwidth]{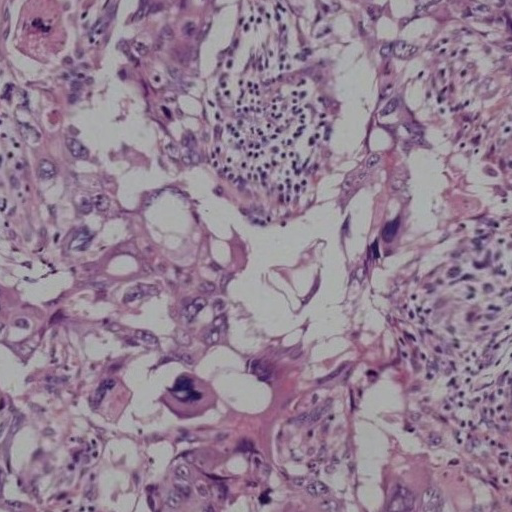} &
        \includegraphics[width=0.15\textwidth]{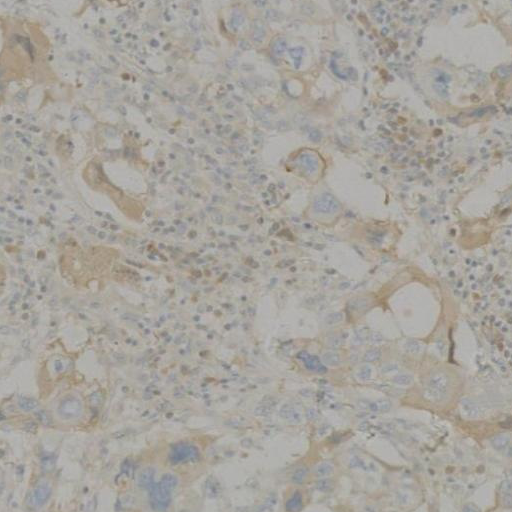} &
        \includegraphics[width=0.15\textwidth]{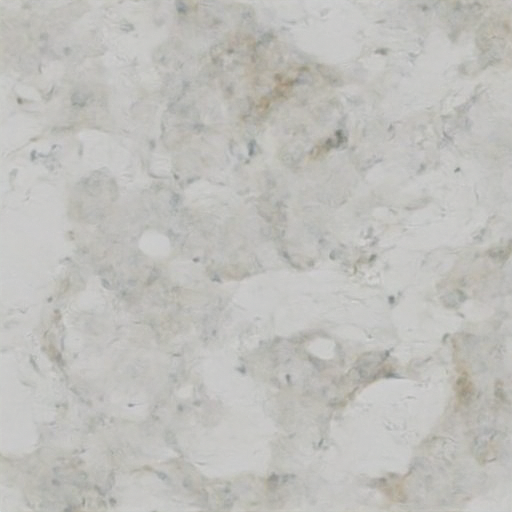} \\
        (a) & (b) & (c) \\
    \end{tabular}
    \caption{Comparison of (a) HE, (b) Ground Truth IHC, and (c) Generated IHC images }
    \label{fig:bci-comparison}
\end{figure}

The results for BCI are reported in Table \ref{tab:bci_comparison}.
Our method performs significantly better in SSIM scores and gives competitive values for PSNR despite being trained on 256x256 and evaluated on 1024x1024 full image. The qualitative results are shown in Fig. \ref{fig:bci-comparison}.
\begin{table}[h]
\centering
\caption{Results on BCI. Test image resolution is 1024$\times$1024. Best method is highlighted in bold font.}
\label{tab:bci_comparison}
\begin{tabular}{l|c|c}
\toprule
Methods                        & PSNR (dB) & SSIM  \\
\midrule
CycleGAN \cite{liu2022bci}     & 16.20   & 0.37 \\
$\mathcal{L}_{\text{ASP}}$ \cite{li2023adaptivesupervisedpatchnceloss}   & 17.86        & 0.50\\
PSPStain \cite{chen2024pathological} &18.62 & 0.45 \\
PPT \cite{zhang2024high}        & 19.09 & 0.49\\
pix2pixHD  \cite{liu2022bci}      & 19.63    & 0.47 \\
BCI \cite{liu2022bci}          & \textbf{21.16}    & 0.47 \\
\midrule
\textbf{Ours}                  & 18.29 & \textbf{0.63} \\
\bottomrule
\end{tabular}
\end{table}

\begin{table*}[!htp] 
\centering
\caption{Comparison with SOTA Methods. Best Method is highlighted in bold font.}
\label{tab:lolv_comparison}
\begin{adjustbox}{max width=\textwidth} 
\begin{tabular}{l|cc|cc|cc|cc}
\toprule
\multirow{2}{*}{Methods} & \multicolumn{2}{c|}{LOL-v1} & \multicolumn{2}{c|}{LOL-v2-real} & \multicolumn{2}{c|}{LOL-v2-synthetic} & \multicolumn{2}{c}{SID} \\
 & PSNR (dB) & SSIM & PSNR (dB) & SSIM & PSNR (dB) & SSIM & PSNR (dB) & SSIM \\
\midrule
SID \cite{chen2018learning}          & 14.35 & 0.43 & 13.24 & 0.44 & 15.04 & 0.61 & 16.97 & 0.59 \\
IPT \cite{chen2021pretrained}                        & 16.27 & 0.50 & 19.80 & 0.81 & 18.30 & 0.81 & 20.53 & 0.56 \\
UFormer \cite{wang2022uformer}                    & 16.36 & 0.77 & 18.82 & 0.77 & 19.66 & 0.87 & 18.54 & 0.57 \\

Sparse \cite{yang2021sparse}                     & 17.20 & 0.64 & 20.06 & 0.81 & 22.05 & 0.90 & 18.68 & 0.60 \\
RUAS \cite{liu2021retinex}                       & 18.23 & 0.72 & 18.37 & 0.72 & 16.55 & 0.65 & 18.44 & 0.58 \\
FIDE \cite{xu2020learning}                       & 18.27 & 0.66 & 16.85 & 0.67 & 23.22 & 0.92 & 19.02 & 0.57 \\

AnlightenDiff \cite{chan2024anlightendiff} 
& 21.72 & 0.81 & 20.65 & 0.83 & - & - & - & - \\

Diff-Retinex \cite{yi2023diff} 
& 21.98 &  0.86 & - & - & - & - & - & - \\

SNR-Net \cite{xu2022snr}                    & 24.61 & 0.84 & 21.48 & 0.84 & 24.14 & 0.92 & 22.87 & 0.62 \\
Retinexformer \cite{retinextransformer}                    & 25.16 & 0.84 & 22.80 & 0.84 & 25.67 & \textbf{0.93} & \textbf{24.44} & \textbf{0.68} \\
PyDiff \cite{zhou2023pyramid}                    &  27.09 & \textbf{0.93} & 24.01 & 0.87 & 19.60 & 0.87 & - & - \\
GSAD \cite{hou2024global}                    & \textbf{27.84} & 0.87 & \textbf{28.82} & \textbf{0.89} & \textbf{28.67} & 0.94 & - & - \\
\midrule
\textbf{Ours}                        & 21.10 & 0.78 & 22.72 & 0.79 & 22.00 & 0.87 & 20.97 & 0.57 \\
\bottomrule
\end{tabular}
\end{adjustbox}
\end{table*}

\subsection{LOL-v1, LOL-v2 and SID Results}
\label{sec:sid_results}
\begin{figure}[h] 
    \centering
    \setlength{\tabcolsep}{1pt} 
    \begin{tabular}{cccc} 
        \includegraphics[width=0.12\textwidth]{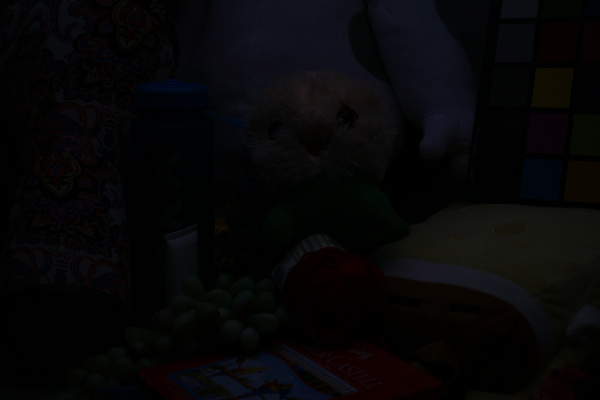} &
        \includegraphics[width=0.12\textwidth]{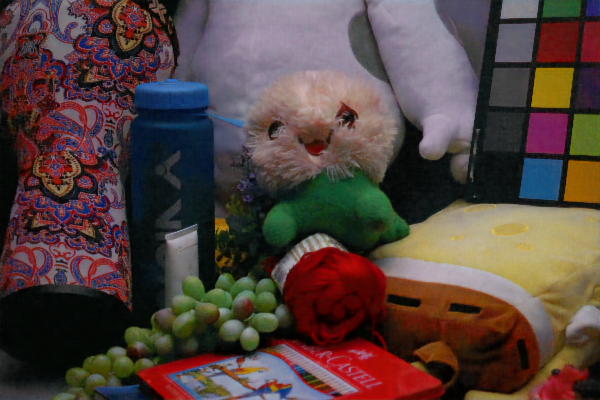} &
        \includegraphics[width=0.12\textwidth]{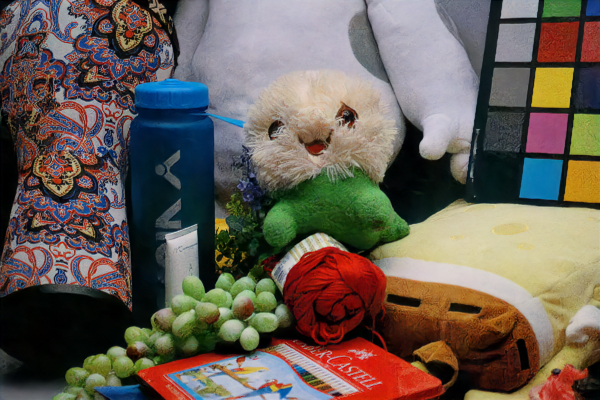} &
        \includegraphics[width=0.12\textwidth]{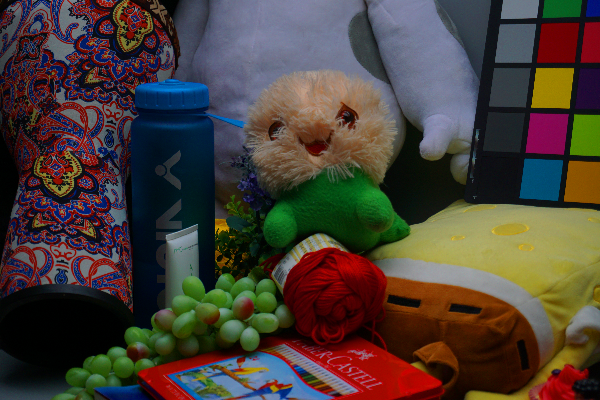} \\
        \includegraphics[width=0.12\textwidth]{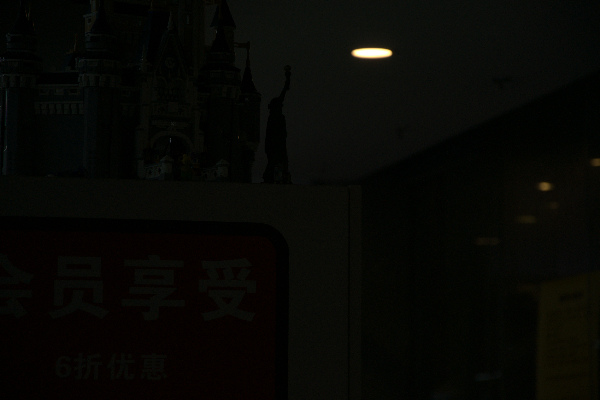} &
        \includegraphics[width=0.12\textwidth]{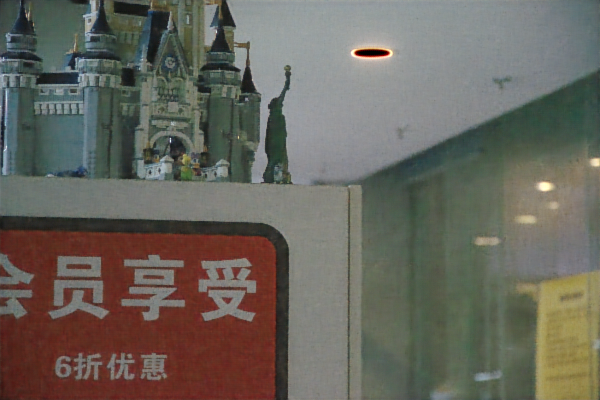} &
        \includegraphics[width=0.12\textwidth]{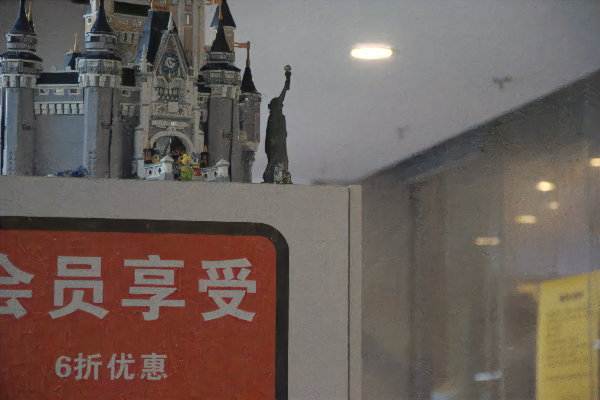} &
        \includegraphics[width=0.12\textwidth]{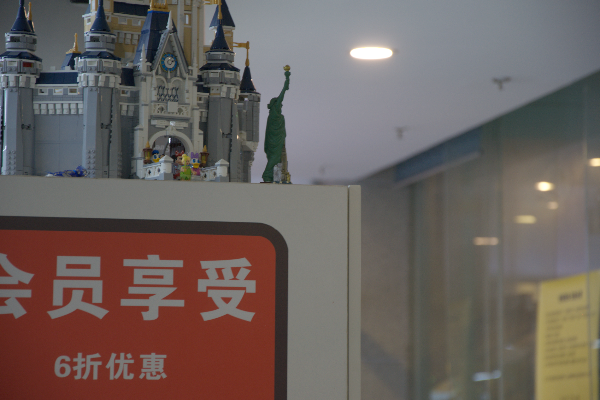} \\
        \includegraphics[width=0.12\textwidth]{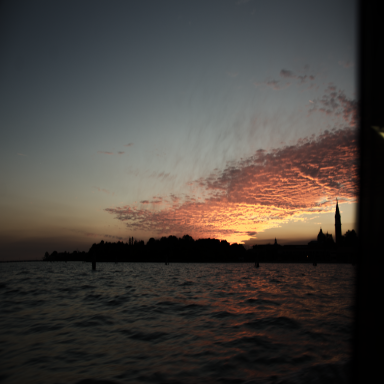} &
        \includegraphics[width=0.12\textwidth]{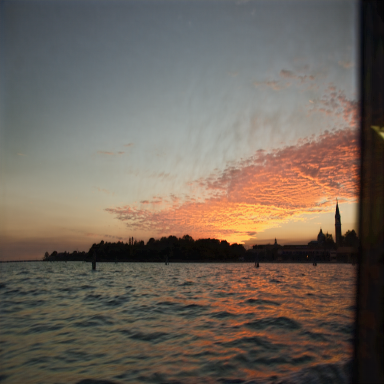} &
        \includegraphics[width=0.12\textwidth]{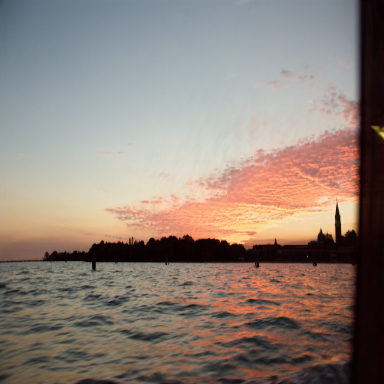} &
        \includegraphics[width=0.12\textwidth]{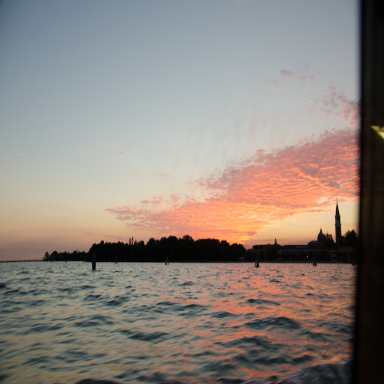} \\
        (a) & (b) & (c) & (d) \\
    \end{tabular}
    \caption{
        Comparison of results on different datasets: 
        Firs row: LoL-v1, Second Row: LoLv2-real, Last row: LoLv2-synthetic. 
        Columns represent: (a) Low-Light Input, (b) RetinexFormer, (c) Ours, and (d) Ground Truth.
    }
    \label{fig:comparison_los}
\end{figure}

We report comparisons for LOLv1, LOLv2-real, and LOLv2-synthetic datasets in Table~\ref{tab:lolv_comparison}. Our method shows a strong performance. However, the results are lower than the SOTA methods. We show the qualitative results in Figure \ref{fig:comparison_los}. We can see that the generated images closely matches to the ground truth.

The results for the Sony subset of the SID dataset \cite{chen2018learning} are shown in Table III. We can see that the proposed method achieves notable performance.

\begin{figure}[!h] 
    \centering
    \setlength{\tabcolsep}{2pt} 
    \renewcommand{\arraystretch}{1.2} 
    \begin{tabular}{ccc} 
        \includegraphics[width=0.15\textwidth]{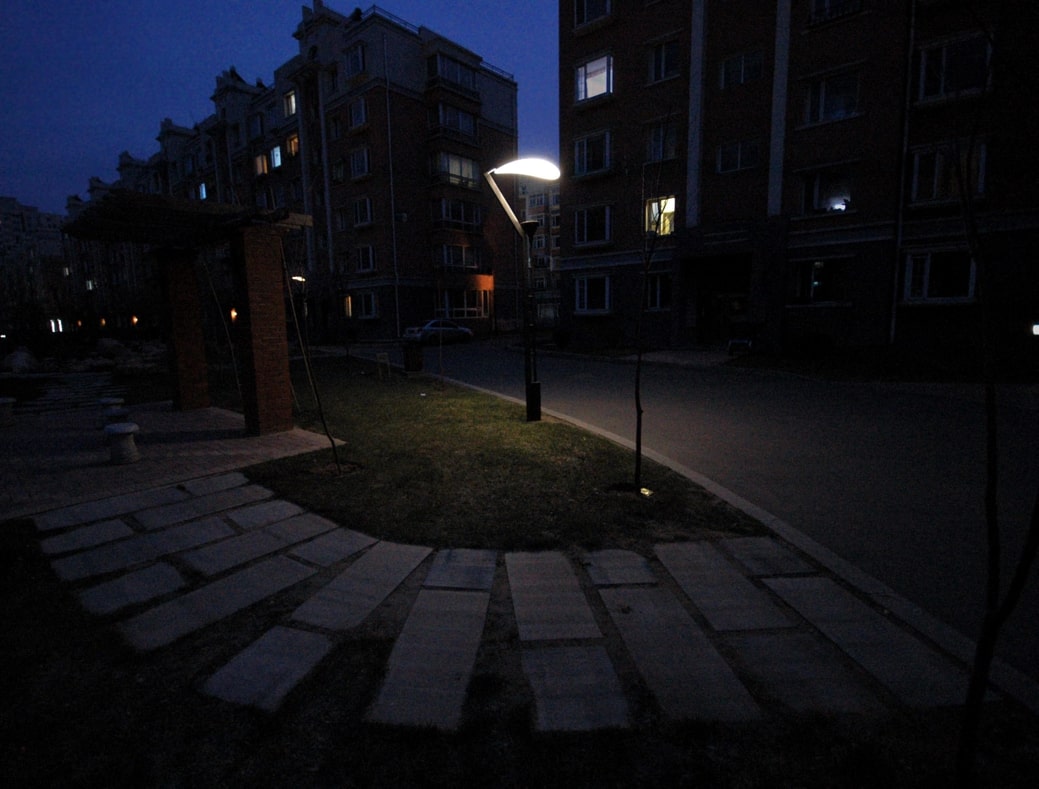} &
        \includegraphics[width=0.15\textwidth]{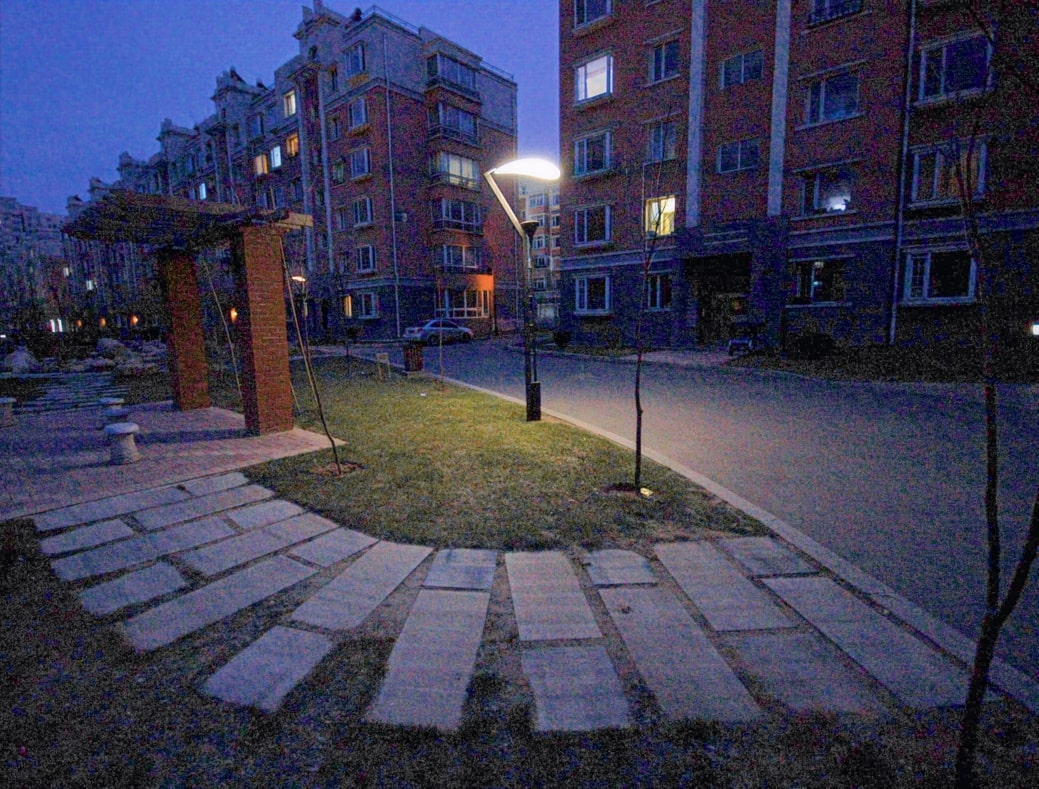} &
        \includegraphics[width=0.15\textwidth]{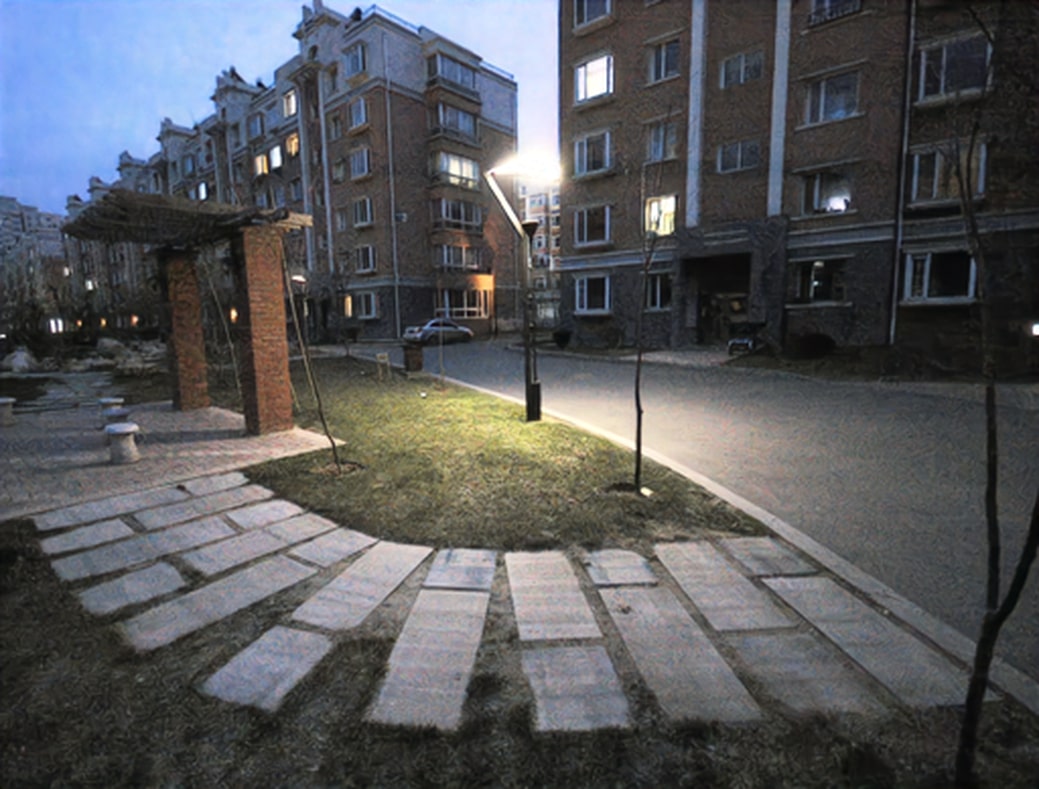} \\

        \includegraphics[width=0.15\textwidth]{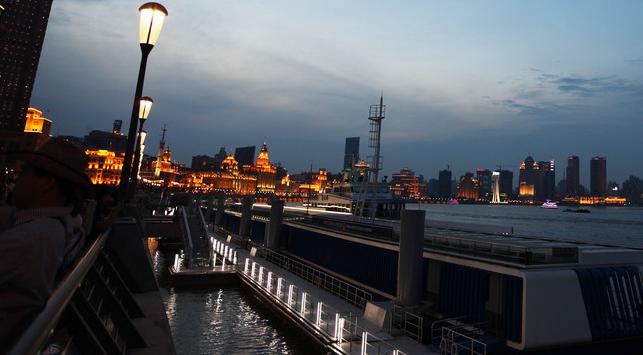} &
        \includegraphics[width=0.15\textwidth]{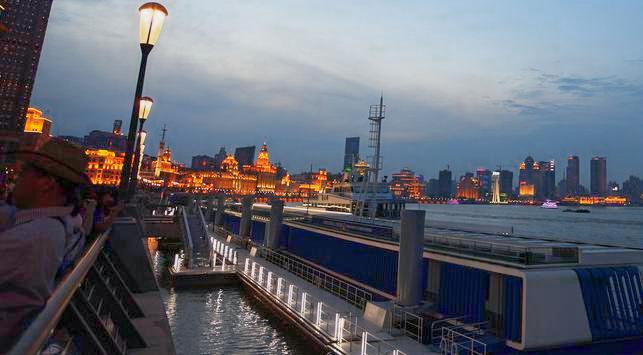} &
        \includegraphics[width=0.15\textwidth]{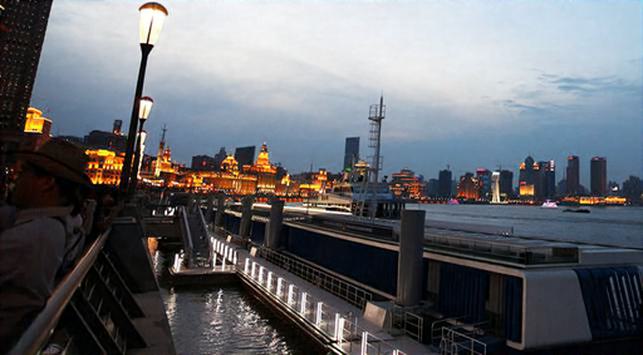} \\

        \includegraphics[width=0.15\textwidth]{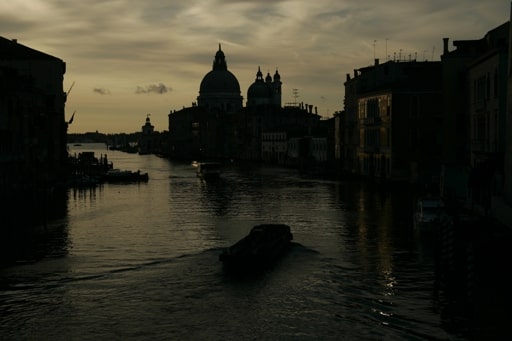} &
        \includegraphics[width=0.15\textwidth]{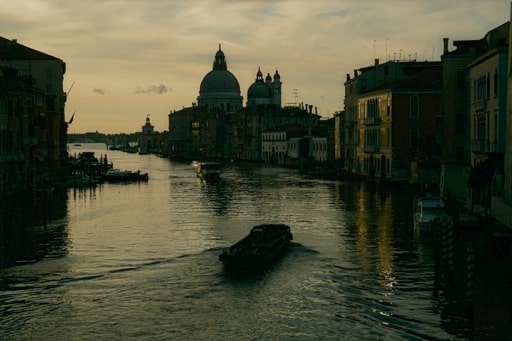} &
        \includegraphics[width=0.15\textwidth]{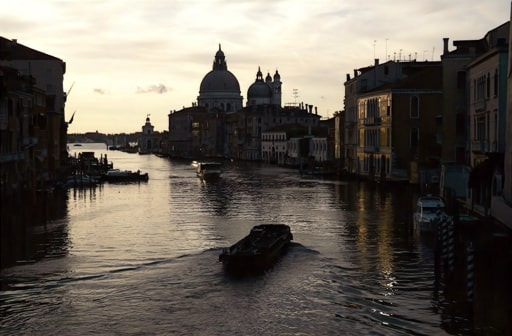} \\

        \includegraphics[width=0.15\textwidth]{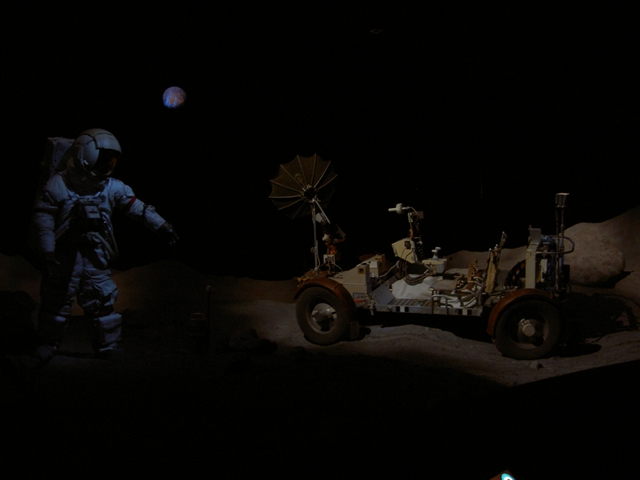} &
        \includegraphics[width=0.15\textwidth]{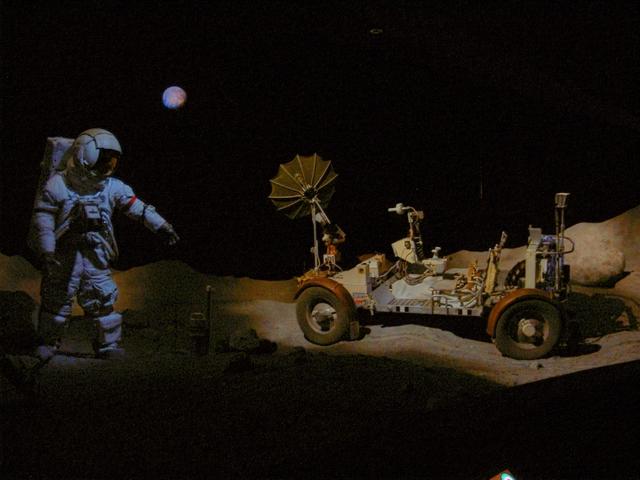} &
        \includegraphics[width=0.15\textwidth]{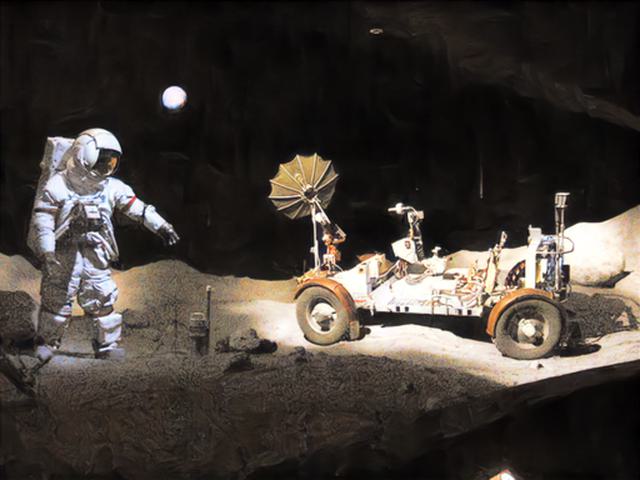} \\

        \includegraphics[width=0.15\textwidth]{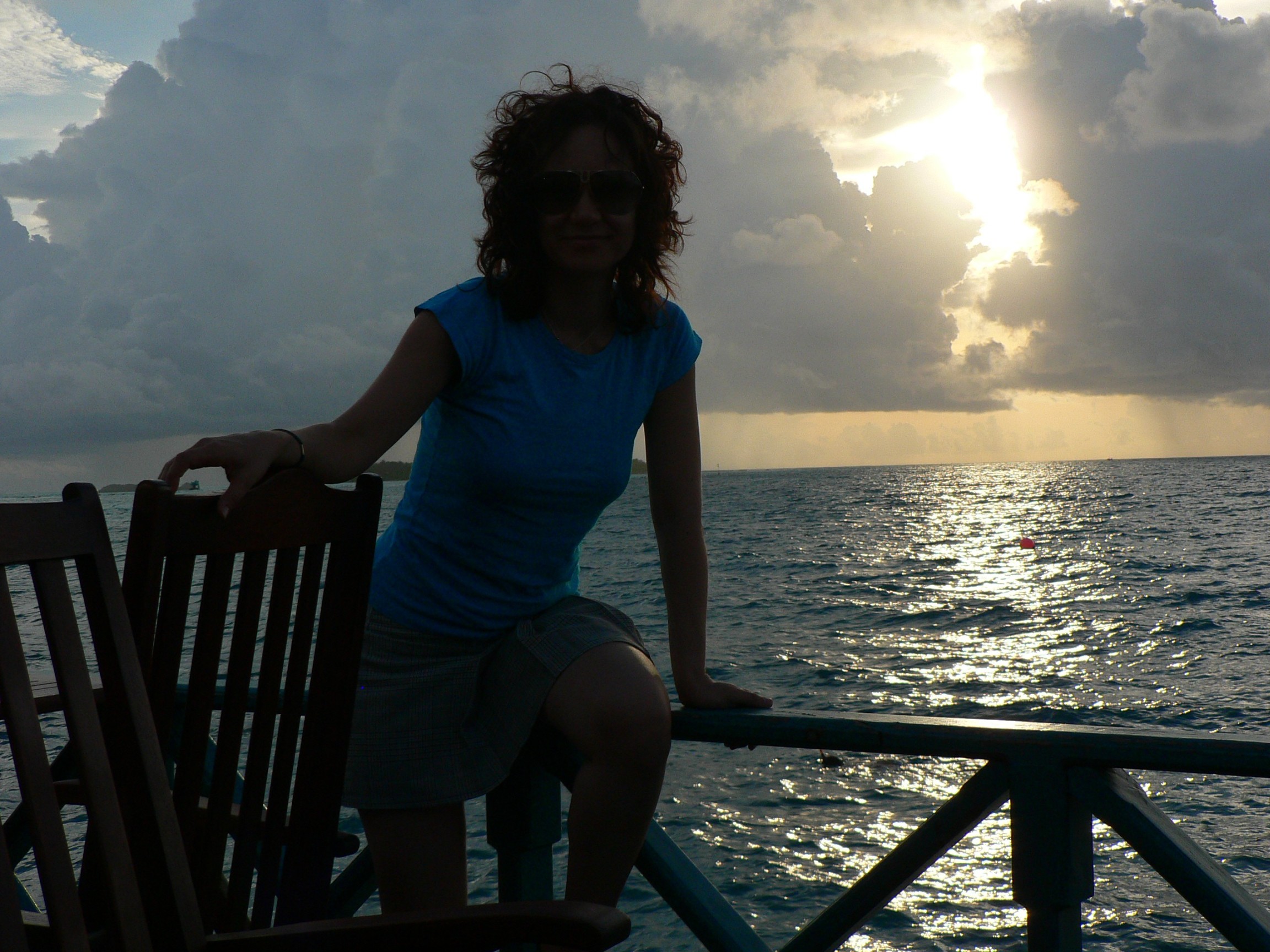} &
        \includegraphics[width=0.15\textwidth]{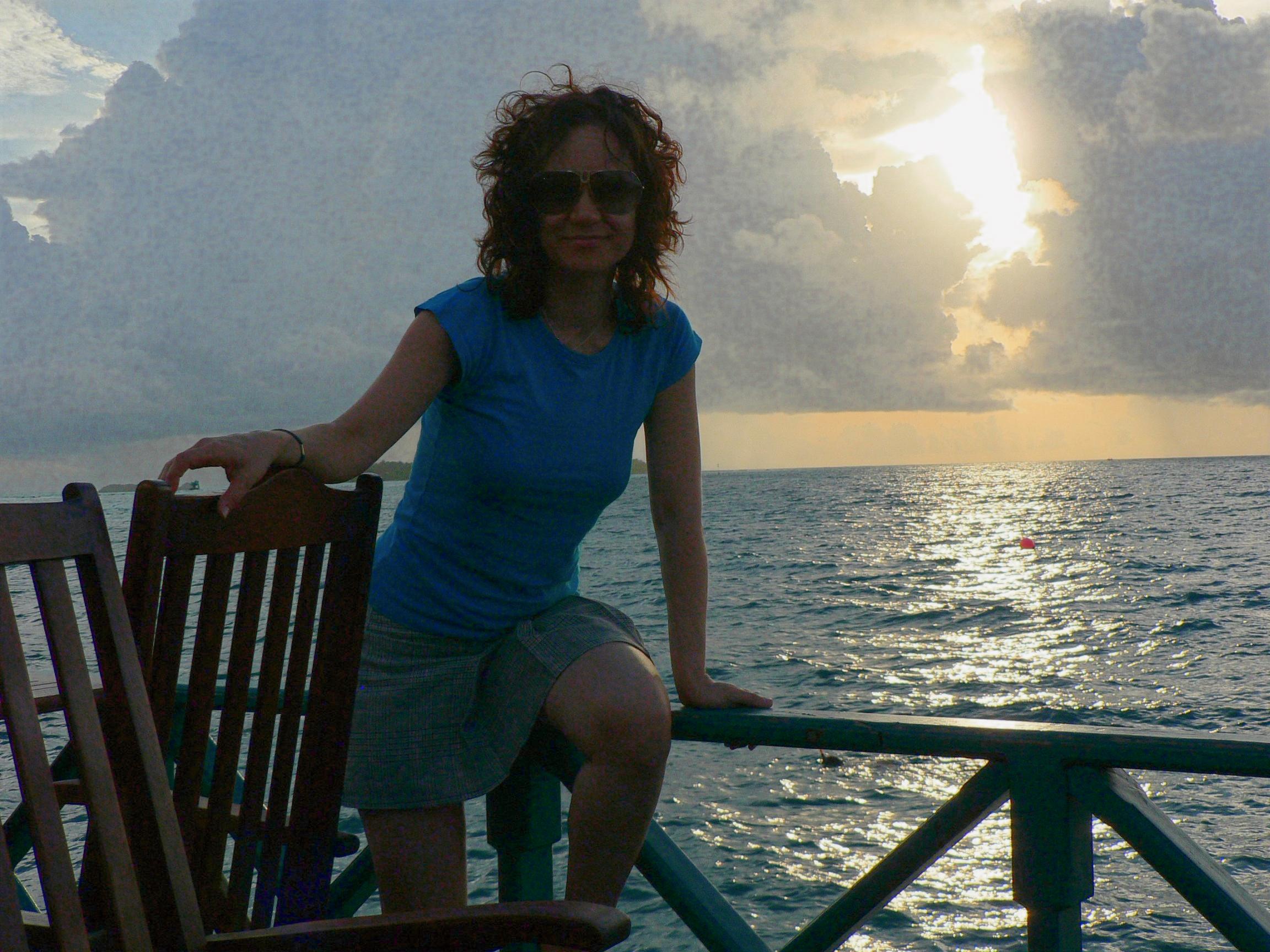} &
        \includegraphics[width=0.15\textwidth]{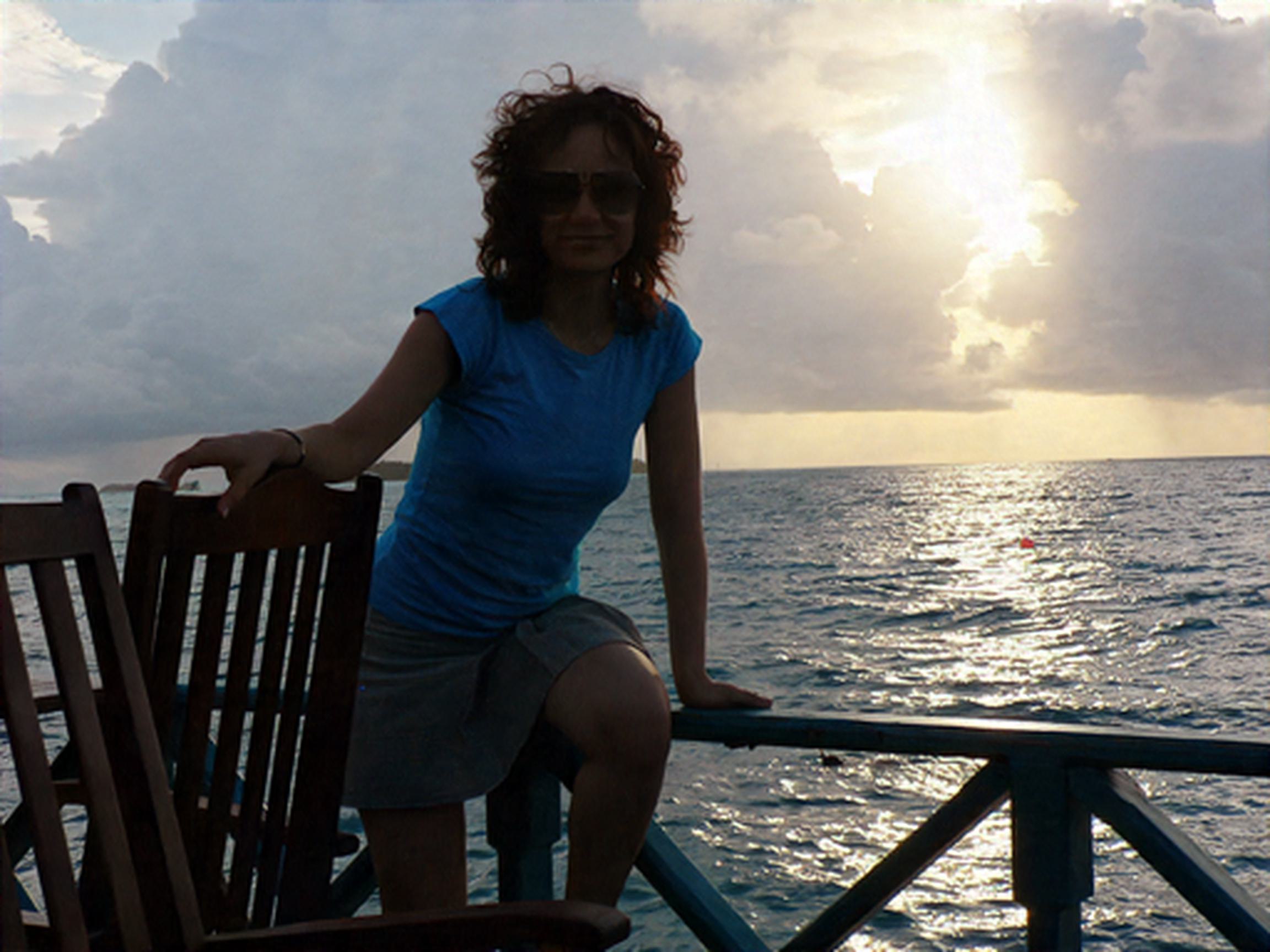} \\
        (a) & (b) & (c)
    \end{tabular}
    \caption{
        First row: LIME, Second row: NPE, Third row: MEF, Fourth row: DICM, Last row: VV. 
        Columns: (a) Input Image, (b) RetinexFormer, (c) Ours.
    }
    \label{fig:comparison_datasets}
\end{figure}

\begin{table}[h]
\centering
\caption{NIQE scores are computed with the full resolution of the images. Lower values indicate better performance. }
\label{tab:niqe}
\begin{tabular}{lccccc}
\toprule
Methods & DICM & LIME & MEF & NPE & VV \\
\midrule
EnGAN \cite{jiang2021enlightengan}  & 3.57 & 3.71 & 3.23 & 4.11 & - \\
KinD \cite{kind} & - & 3.88 & 3.34 & 3.92 & - \\
DCC-Net \cite{dccnet2022}  & 3.70 & 4.42 & 4.59 & 3.70 & 3.28 \\
GSAD \cite{hou2024global}  & 3.28 & 4.32 & 3.40 & \textbf{3.55} & 2.60\\
Retinexformer \cite{retinextransformer} & \textbf{2.85} & 3.70 & 3.14 & 3.64 & \textbf{2.55} \\
\midrule
 \textbf{Ours} & 3.09 & \textbf{3.67} & \textbf{2.96} & 3.65 & 4.7\\
\bottomrule
\end{tabular}
\end{table}

\subsection{NIQE Scores}
\label{sec:niqe_results}
We compare the naturalness image quality (NIQE) across DICM, LIME, MEF, NPE, and VV datasets in Table \ref{tab:niqe}. Our model achieves the best results for LIME and MEF, and the second-best results for DICM. As the model does not see these datasets during training, better NIQE scores indicate stronger generalization to unseen domains. For VV, a high-resolution dataset, using a model trained on LoLv2-synthetic with 128x128 patches causes a performance drop. 

Compared to SOTA method GSAD \cite{hou2024global} in LOL-v1 and LOL-v2, we perform better in DICM, LIME and MEF. In case of VV, \cite{hou2024global} performs better. Compared to SOTA method Retinexformer \cite{retinextransformer} in SID, our method does better for LIME and MEF, and gives approximately same performance in NPE. The qualitative results in Fig. \ref{fig:comparison_datasets} clearly show enhanced details.

\section{Conclusion}
In this work, we introduced Conditional Consistency Models (CCMs) for cross-modal image translation and enhancement tasks. The conditional input guides the denoising process and generates output corresponding to the paired conditional input. Unlike existing methods such as GANs and diffusion models, CCMs achieve superior results without requiring iterative sampling or adversarial training. Distinct from other works, our method can be adopted for different tasks of translation or enhancement and shows competitive results in both the tasks.
Extensive experiments on benchmark datasets demonstrate the superior performance of our method. 
In future, we aim to explore further generalization of CCMs to additional conditional tasks and investigate improvements in conditional guidance mechanisms.

\small 
\bibliographystyle{IEEEbib}
\bibliography{icme2025references}
\end{document}